\documentclass[preprint,12pt]{elsarticle}

\usepackage{amssymb}
\usepackage{amsmath}
\usepackage{algorithm}
\usepackage{algpseudocode}
\usepackage{multirow}
\usepackage{lscape}

\journal{Expert Systems with Applications}

\begin{document}

\begin{frontmatter}

\title{Rumour detection using graph neural network and oversampling in benchmark Twitter dataset}

\author[inst1]{Shaswat Patel}
\author[inst1]{Prince Bansal}
\author[inst1]{Preeti Kaur}

\affiliation[inst1]{organization={Department of Computer Engineering},
            addressline={Netaji Subhas University of Technology}, 
            city={New Delhi},
            postcode={110078}, 
            state={New Delhi},
            country={India}}

\begin{abstract}
Recently, online social media has become a primary source for new information and misinformation or rumours. In the absence of an automatic rumour detection system the propagation of rumours has increased manifold leading to serious societal damages. In this work, we propose a novel method for building automatic rumour detection system by focusing on oversampling to alleviating the fundamental challenges of class imbalance in rumour detection task. Our oversampling method relies on contextualised data augmentation to generate synthetic samples for underrepresented classes in the dataset. The key idea exploits selection of tweets in a thread for augmentation which can be achieved by introducing a non-random selection criteria to focus the augmentation process on relevant tweets. Furthermore, we propose two graph neural networks(GNN) to model non-linear conversations on a thread. To enhance the tweet representations in our method we employed a custom feature selection technique based on state-of-the-art BERTweet model. Experiments of three publicly available datasets confirm that 1) our GNN models outperform the the current state-of-the-art classifiers by more than 20\%(F1-score); 2) our oversampling technique increases the model performance by more than 9\%;(F1-score) 3) focusing on relevant tweets for data augmentation via non-random selection criteria can further improve the results; and 4) our method has superior capabilities to detect rumours at very early stage.
\end{abstract}

\begin{keyword}
Oversampling \sep Data augmentation \sep Graph Neural Network \sep BERT
\end{keyword}

\end{frontmatter}


\section{Introduction}
\label{sec:introduction}
Over the last decade, internet usage has increased manifold leading to substantial growth in usage of social media platform. On Twitter alone 456,000 Tweets are sent every minute of the day.\cite{twitterDataUsage} Social media platforms like Twitter have become a prominent source for people to gather news apart from traditional media outlets.\cite{twitterNews} Due to the expeditious and unmoderated nature of these platforms  it  has unfortunately become a breeding ground for rumours. The rumours on social media sites have become a nuisance for the users which has valiantly transferred power to mislead people.\cite{rumourMislead} In the 2014 and 2016 U.S. presidential elections the online social media sites like Facebook and Twitter were widely used to spread both verified and rumour for and against various candidates\cite{2016Election, 2012Election}. False claims decisively moulds people's perception over events which is often harmful. Due to the increasing reliance on social media for news updates, the disruptive impact of rumours is further accentuated. 

Defining rumour is a crucial and critical task, while in the literature there are various definitions being followed. According to the Oxford Dictionary, rumour can be defined as follows: \emph{Information or a story that is passed from one person to another and may or may not be true.} This definition is consistent with majority of the works in the literature.\cite{rumourDetectionLotfi, zubiaga2015detecting, Zubiaga2019, 10.1007/978-981-16-2597-8_8} However, multiple works exists which have considered rumour as false information.\cite{CaiRumourDetection, 10.1145/2350190.2350203} In our paper, a rumour is a piece of information that is unverified or not vetted at the time of reporting. Rumour detection in the context of Twitter is classification of tweets into three categories true, false or unverified.

Notably, rumor detection is a challenging task due to the following three aspects: 1) The characteristics of rumours are often delusional as they are intentionally well designed to imitate real news for various motivations, such as political astroturfing, conducting malicious marketing management. It is difficult for ordinary people or even domain experts to distinguish between true rumors and false rumors. 2) The complex interaction between various user on a rumor spreading tweet is often difficult to interpret. Users on social networks are particularly active, which enables a variety of information to spread widely in a short time.  3) Lastly, the disproportionate nature of data available for various labels and events, make it difficult to create robust and highly generalizable models.

In this paper, we propose a novel method for rumour detection which utilises user comments and simultaneously handle the scarcity of data across different events and labels. It has been observed that users on social media sites collectively verify a piece of information through comments.\cite{3162956} A thread on Twitter is a series of connected Tweets. We represent a twitter thread as graph with nodes which consists of source tweet and user comments. We have employed Graph Neural Networks(GNN) to translate this spatiality of a Twitter thread into a graph embedding. Along with this, we have incorporated textual and user behavioral features. To increase the generalizability and reduce overfitting while training our method, we propose a novel strategy for oversampling the dataset using data augmentation. Data Augmentation aims at generating synthetic training dataset in scenarios where data is insufficient.\cite{li2021data}  Distillation of tweets for data augmentation was done based on influence score which is co-related to number of words in a tweet. The influence score in combination with contextualized data augmentation method\cite{ma2019nlpaug} provides the necessary basis for augmenting a tweet thread.

We have organized our paper into following section: Section \ref{sec2} discusses the background work done within the domain of rumour detection. Section \ref{sec3} formally describes the problem statement with essential definitions. Section \ref{sec4} discusses the proposed oversampling technique followed by Section \ref{sec5} detailing our proposed method. Section \ref{sec6} discusses the results followed by conclusion of the empirical study in Section \ref{sec7}.

\section{Background work}\label{sec2}
Detection of rumour on social media site is essential, keeping in mind the volume of User-generated data. Currently, most of the rumour detection techniques are supervised. In 2017, Enayet et al.\cite{enayet-el-beltagy-2017-niletmrg} proposed NileTMRG where-in they used various preprocessing, feature extraction and selection, and learning classifiers. The linear support vector machine classifier(Linear SVC) had the best accuracy during cross-validation. In 2018, Kochkina et al.\cite{Kochkina2018AllinoneML} proposed a multi-task learning method for rumour detection on PHEME and RumourEval dataset. They assessed four multi-task learning approaches: (1) single task learning to perform veracity classification, (2) multi-task learning that combines stance and veracity classification, (3) multi-task learning that combines rumour detection and verification, and (4) performing multi-task learning that combines rumour detection, stance classification and veracity prediction. They determinted that multi-task learning on two tasks greatly outperforms single task learning. Additionally, multi-task learning on all three task further improves the results. Ma et al.\cite{ma-etal-2018-rumor} proposed a RvNN based approach which considers content semantics and propagation cues trained on Twitter15 and Twitter16 dataset. They extended the RvNN architecture into two variants: Bottom-up(BU) and Top-down(TD) model to represent the propagation structure. In 2019, Asghar et al.\cite{Asghar2019} proposed BiLSTM-CNN trained on PHEME dataset. BiLSTM-CNN model uses BiLSTM layers to learn long term dependency in a tweet, followed by CNN layers to extract features for accurate prediction. Wei et al.\cite{wei-etal-2019-modeling} proposed a a hierarchical multi-task learning framework named Hierarchical-PSV for jointly predicting rumor stance and veracity on PHEME and SemEval-2017 task 8 dataset. They exploited both structural and temporal aspect of rumour detection to predict user stance through Conversational-GCN and veracity through Stance-Aware RNN. In 2020, Bain et al.\cite{https://doi.org/10.48550/arxiv.2001.06362} proposed a Bi-directional GCN that uses both bottom-up and top-down propagation captured by bottom-up graph convolution network and top-down graph convolution network respectively. Furthermore, to avoid overfitting they employed DropEdge in training phase. To show effectiveness of their model they trained the model on three datasets, Weibo, Twitter15 and Twitter16. In 2021, Kumar et al.\cite{Kumar2021} proposed $CNN_{IG-ACO}NB$ model trained on PHEME dataset. The proposed model, employs CNN which acts as a feature learner and Naive Bayes as a classifier. The Naive Bayes is trained on CNN generated features which are further optimized by Information gain—Ant colony Optimizer(IG-ACO). Wei et al.\cite{wei-etal-2021-towards} proposed Edge enhanced Bayesian Graph Convolutional Network
(EBGCN) model to handle uncertainties in propagation structure using Bayesian method by adjusting weights of unreliable relations. Additionally, they proposed an unsupervised edge-wise consistency training framework to learn latent relations. To validate their work, they used three real world dataset, PHEME, Twitter15 and Twitter16.

\begin{table}[H]
\begin{tabular}{|p{0.75cm}|p{2.75cm}|p{1.75cm}|p{2cm}|p{3.8cm}|}
\hline
Year &
  Technique Applied &
  Dataset &
  Feature Set &
  Conclusion \\ \hline
2021 &
  EBGCN &
  Twitter15, Twitter16, PHEME &
  Message passing based &
  EBGCN significantly outperforms baselines on both rumor detection and early rumor detection tasks \\ \hline
2021 &
  CNN, IG-ACO &
  PHEME &
  Textual features combined with feature vector &
  Results validate superior F1 of 0.732 using the proposed CNN+IG-ACONB rumour classifer \\ \hline
2020 &
  Bi-GCN &
  Weibo, Twitter15, Twitter16 &
  Propagate Structure based on retweet and response relationships &
  The BiGCN model achieves the best performance by considering both the causal features of rumor propagation along relationship chains from top to down propagation pattern \\ \hline
2019 &
  Hierarchical-PSV, Conversational-GCN,  Stance-Aware RNN &
  SemEval-2017 task-8, PHEME &
  Perspective based,  learning tweet representations through aggregating information from neighboring tweets. &
  Conversational-GCN can handle deep conversation structures effectively and hierarchical framework performs much better than existing methods \\ \hline
\end{tabular}
\end{table}

\begin{table}[H]
\begin{tabular}{|p{0.75cm}|p{2.75cm}|p{1.75cm}|p{2cm}|p{3.8cm}|}
\hline
Year &
  Technique Applied &
  Dataset &
  Feature Set &
  Conclusion \\ \hline
2019 &
  BiLSTM-CNN &
  PHEME &
  Long term dependency in a tweet by considering both the previous (past) and next (future) context information. &
  BiLSTM-CNN performed better than the other models with accuracy of 86.12\% \\ \hline
2018 &
  Tree-structured Recursive Neural Networks &
  Twitter15, Twitter 16 &
  Propagation layout of tweets &
  Results on two public Twitter datasets show that given method improves rumor detection performance in very large margins as compared to baselines. \\ \hline
2018 &
  Multi-task learning, LSTM &
  SemEval, PHEME &
  Joint learning of several related tasks with a shared representation, parameter sharing &
  The joint learning of two tasks from the verification pipeline outperforms a single learning approach to rumour verification. Combination of all three tasks leads to further performance improvement \\ \hline
2017 &
  Linear SVC &
  SemEval RumourEval &
  User and Content based &
  The identity and behavior of the user didn’t affect much the credibility of the rumour he/she is spreading. \\ \hline
\end{tabular}
\caption{Related work with technique applied, dataset, feature set used and conclusion.}
\end{table}

\section{Notations and Problem Statement}\label{sec3}

In this section, we first introduce some necessary definitions that were used throughout the paper. Later, we formally define the problem statement of rumour detection task.

\textbf{Source tweet}: A source tweet is the tweet that starts the thread and is not a reply to any other tweet. It is denoted by $s_i$ where i denotes the thread the source belongs.

\textbf{Response tweets or comments}: Tweets that are replies to the source tweet or other responsive tweet showcasing the thoughts of a user on the topic being discussed. It is denoted by $c^i_j$ where i denotes the Twitter thread the response belongs while the j denotes the order of response with respect to the time at which it was posted online. 

\textbf{Twitter thread}: A Twitter thread \emph{t} is a set of related tweets, \(M = \{s_1, c_1, ..., c_n\}\) where $s_1$ is the source tweet while $c_1, ..., c_n$ are the responses received on the source tweet. 

\begin{figure}
    \includegraphics[width=\textwidth]{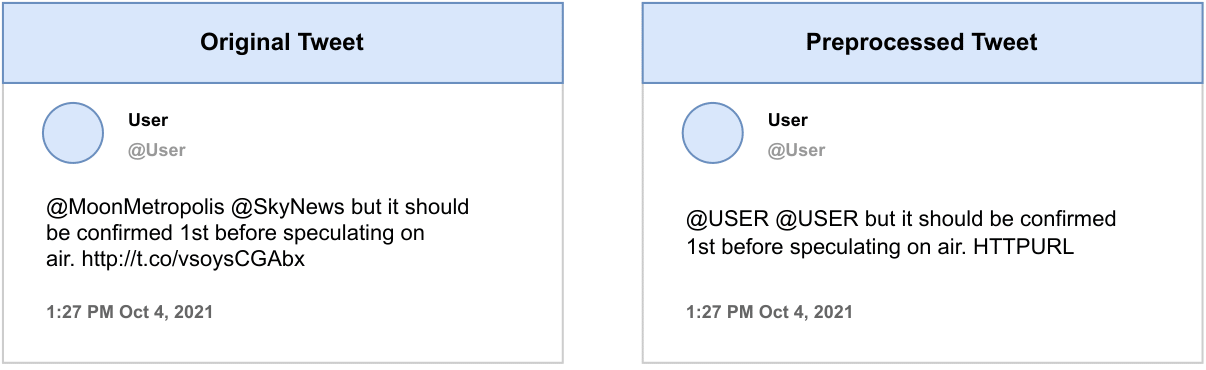}
    \centering
    \caption{An example of Twitter thread and Propagation Graph}
    \label{fig:twitterThread}
\end{figure}

A Twitter thread can be represented as a graph. This graph is also called the propagation graph as it shows the flow of information from the source tweet to any other tweet in the Twitter thread. Fig \ref{fig:twitterThread}(a), shows the the Twitter thread while Fig \ref{fig:twitterThread}(b) shows the corresponding propagation graph with the root of the graph as the source tweet followed by the responses in the thread.

Each node in the graph has two relationships, in-response and responses-on as observed in Fig \ref{fig:twitterThread}(c). The two relationships are in-parallel with parent and child relation usually observed in a graph structure. Each tweet in a thread other than the source tweet is in-response of some other tweet and each tweet other than the leaves of the graph have responses to them(response-on).

\textbf{Rumour detection}: The rumour detection problem can be considered as a  Multi-classification task. 
For each tweet we have multiple properties representing its content, such as text and image. Each tweet is also associated with a user, each user also has a set of properties, including name, number of followers, verified etc. A Twitter thread also contains \emph{responsive tweets} or \emph{comments}. The \emph{rumour detection} task is then defined as: Given a Twitter thread \emph{t} the rumor detection task aims to determine whether this story is true, false or unverified. This definition formulates the rumor detection task as a veracity classification task. This definition is consistently followed in many studies\cite{https://doi.org/10.48550/arxiv.1807.03505, https://doi.org/10.48550/arxiv.1712.07709, 10.5555/3061053.3061153, Zhou_2020, https://doi.org/10.48550/arxiv.1911.07199}.

Therefore, our aim is to train an accurate classifier,$f$, which uses the Twitter thread, that is $f:t^{(j)} \rightarrow Y^{(j)}$, where $Y^{(j)}$ is a multi-classification label, given the Twitter thread $t^{(j)}$.

\section{Multifold Oversampling of Tweet Thread}\label{sec4}
%

    
    
    
    
    
    


%

A very common real-world challenge for training various machine learning and deep leaning models is the unavailability of balanced datasets. To alleviate this issue, various oversampling techniques has been proposed for removing the skewness across datasets. Random re-sampling is a very common approach to handle such data disparity. This technique can be accomplished in two ways either by randomly duplicating some examples of the minority class, called oversampling or by randomly deleting samples from the majority class, called undersampling. In real-world datasets the number of samples available for training a model is significantly less, along with this the disparity in the dataset is quite large. Hence, reducing the dataset using undersampling is generally not the most effective method\cite{Buda_2018}.

Recently, to diversify the oversampled data, data augmentation techniques merged with oversampling has been replicated on plethora tasks(QA task, Image captioning, etc.).\cite{Buda_2018, 10.1007/978-981-16-2594-7_38, gosain_handling_2017} Data augmentation entails generation of new data points by slight modification of existing data points. There are various techniques such as synonym based, semantic word embedding based, and character error based. These techniques are not able to fully grasp the lingo used on Twitter. To capture the diversity of linguistics of Twitter, language models are trained on large scale Twitter corpora. BERTweet is one such language model which is trained on 850 million tweets\cite{bertweet} which we have leveraged for data augmentation using the NLPAUG library\cite{ma2019nlpaug}.

The benchmark datasets available for rumour detection are imbalanced in nature(Table \ref{table:dataDistribution}) necessitating in utilization of oversampling. We propose a novel oversampling technique, Multi-fold Oversampling(MOS), which generates new tweet threads from existing dataset using contextualised data augmentation of source and response tweets. Since the rumour dataset is classified based on events, MOS works on individual events which helps in generating tweets for new events with high precision. 

\begin{figure}
    \includegraphics[width=\textwidth]{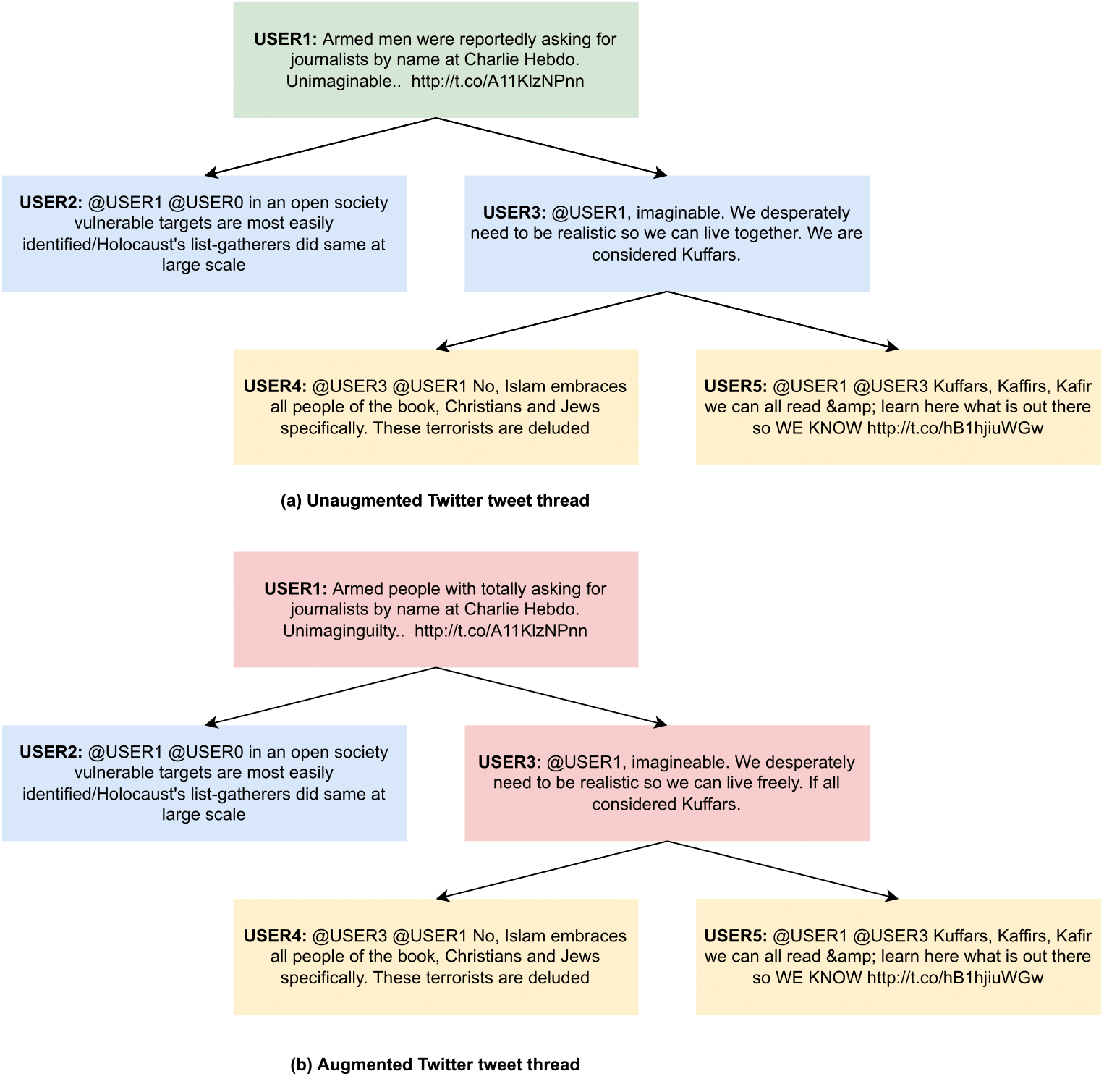}
    \centering
    \caption{Example of contextualized data augmentation in MOS}
    \label{fig:dataAug}
\end{figure}

MOS combines the benefits of both oversampling and data augmentation to generate robust training examples(Figure \ref{fig:dataAug}). In MOS, each event is balanced by oversampling tweet threads for underrepresented classes. Oversampling in MOS can be employed by using random or non-random data augmentation techniques. The primary algorithm for MOS is present in algorithm \ref{algorithm:dataaugalgo} where each event is handled independently. Maximum number of tweets in each event is calculated initially and underrepresented classes are oversampled to evenly distribute each event by defining the $OVERSAMPLE$ function. 

\subsection{Random data augmentation based Oversampling}
Data augmentation offers the ability to generate synthetic data using the available data distribution.\cite{10.1007/978-981-16-2594-7_38} In random data augmentation based oversampling (R-MOS) technique, a tweet thread \emph{T} is selected which will be augmented \emph{n} times. Each \emph{T} constitute multiple tweets which are selected randomly for augmentation using nlpaug context based augmentation technique which uses a language model to either insert or substitute words in the text using surrounding context. We have used BERTweet as the language model to substitute words in the tweet as BERTweet understands the context of a tweet due its the ability to capture the social media lingo. Also, BERTweet is able to capture the bi-directional context for a word being substituted.\cite{bertweet} 

For each category we initially calculate $n_{random}$ and $n_{fold}$ where $n_{random}$ is the number of threads to select at a random and $n_{fold}$ is the number of times we augment the entire set of threads. For selection purposes, we define $p_{aug}$ which is the percentage of tweets in \emph{T} to be augmented which we have set as 20\%. Tweets which are augmented are then replaced in the thread to get $thread_{augmented}$.

\subsection{Non-random data augmentation based Oversampling}
One drawback of R-MOS is that each tweet is given equal importance for data augmentation. As observed in Appendix Figure \ref{fig:relevancy}. not all comments hold equal importance as few comments contain only images or emojis. Tweets with only images or emojis cannot be augmented as it will act as noise for the classifier. Based on this observation we propose a hypothesis for the selection of relevant tweets for data augmentation. 

Non-random data augmentation(NR-MOS) utilizes number of words as pivotal parameter to determine the tweet relevancy from a given Twitter thread. Higher the number of words in the tweets, better the semantics of contextualized embeddings generated from those tweets.  Based on the proposed hypothesis a non-random selection criteria, \emph{influence score}, is used for selection of responses from a tweet thread(Algorithm \ref{algorithm:nraugalgo}). The influence score in Equation \ref{equation:influencescore} is calculated by the difference between number of words in the tweet and number of keywords present in the tweet. All special assigned words or keywords(e.g. HTTPURL, @USER) as mentioned in data preprocessing section. 

\begin{equation}
\textit{Influence Score}(c_i) = \textit{Number of words}(c_i) - \textit{Number of keywords}(c_i) \forall i \in t
\label{equation:influencescore}
\end{equation}

A probability distribution is generated using the influence score which is used to facilitate the selection of tweets during the process of augmentation for tweet thread.

\begin{algorithm}[H]
\caption{Twitter thread data augmentation}\label{algorithm:dataaugalgo}
\begin{algorithmic}
\For{each EVENT e in DATASET}
    \State $t_{label} \in e$
    \State $n_{label} \leftarrow \|t_{label}\|$
    \State $n_{max} \leftarrow max(n_{label})$
    \For{each l $\in$ label}
        \State $OT_l \leftarrow t_l$ \Comment{OT is set of oversampled tweet threads}
        \If{$n_l \neq 0$}
            \State $OT_{oversampled} \leftarrow OVERSAMPLE(t_l, n_{max} - n_l)$
            \State $OT_l \leftarrow t_l \cup OT_{oversampled}$
        \EndIf
        \State $e' \leftarrow e' \cup OT_l$
    \EndFor
\EndFor
\end{algorithmic}
\end{algorithm}



\begin{algorithm}[H]
\caption{Random augmentation algorithm}\label{algorithm:raugalgo}
\begin{algorithmic}

\Function{RANDOM\_OVERSAMPLE}{$t_{label}$, n}
\State $n_{label} \leftarrow \|t_{label}\|$
\State $n_{random} \leftarrow n \% n_{label}$
\State $tweet_{random} \leftarrow random(t_{label}, n_{random})$
\State $n_{fold} \leftarrow min(n / n_{label}, 3)$
\State $ OT \leftarrow {\phi} $
\For{$t_i$ in $t_{label}$}
    \State $OT \leftarrow OT \cup RANDOM\_AUGMENT(t_i, n_{fold})$
    \If{$ t_i $ in $tweet_{random}$}
      \State $OT \leftarrow OT \cup RANDOM\_AUGMENT(t_i, 1)$
    \EndIf
\EndFor
\State RETURN OT
\EndFunction
\State
\State
\Function{RANDOM\_AUGMENTATION}{T, n}
\State $result \leftarrow \{\phi\}$
\For{$i \in \{1, 2,..., n\}$}
    \State $tweet_{index} \leftarrow random(T, p_{aug} * \|T\|)$ 
    \State $thread_{augmented} \leftarrow T$
    \For{$t_i$ in $tweet_{index}$}
        \State $tweet_{augmented} \leftarrow nlpaug( tweet[t_i] ) $
        \State $thread_{augmented}[t_i] \leftarrow tweet_{augmented}$
    \EndFor
    \State $result \leftarrow result \cup thread_{augmented}$
\EndFor
\State RETURN result
\EndFunction
\end{algorithmic}
\end{algorithm}

\begin{algorithm}[H]
\caption{Non-random augmentation algorithm}\label{algorithm:nraugalgo}
\begin{algorithmic}

\Function{NONRANDOM\_OVERSAMPLE}{$t_{label}$, n}
\State $n_{label} \leftarrow \|t_{label}\|$, $n_{random} \leftarrow n \% n_{label}$
\State $t_{random} \leftarrow random(t_{label}, n_{random})$, $n_{fold} \leftarrow min(n / n_{label}, 3)$
\State $ OT \leftarrow \{\phi\} $
\For{$t_i$ in $t_{label}$}
    \State $OT \leftarrow OT \cup NONRANDOM\_AUGMENT(t_i, n_{fold})$
    \If{$ t_i $ in $t_{random}$}
      \State $OT \leftarrow OT \cup NONRANDOM\_AUGMENT(t_i, 1)$
    \EndIf
\EndFor
\State RETURN OT
\EndFunction
\Function{GET\_PROBABILITY}{$T$}
\State PROBABILITY $\leftarrow$ \{$\phi$\}
\For{tweet in T}
    \State $num_{words} \leftarrow 0 $
    \For{word in $T$}
        \State $num_{words} \leftarrow num_{words} + 1$
    \EndFor
    \State $num_{keywords} \leftarrow 0$
    \For{word in $T$}
        \If{ word in keywords }
            \State $num_{keywords} \leftarrow num_{keywords} + 1$
        \EndIf
    \EndFor
    \State $weight \leftarrow num_{words} - num_{keywords}$
    \State PROBABILITY $\leftarrow$ PROBABILITY $\cup$ weight 
\EndFor
\State RETURN PROBABILITY
\EndFunction
\Function{NONRANDOM\_AUGMENTATION}{T, n}
\State $result \leftarrow \{\phi\}$
\For{$i \in \{1, 2,..., n\}$}
    \State probability $\leftarrow$ GET\_PROBABILITY(T)
    \State $tweet_{index} \leftarrow random(T, p_{aug} * \|T\|, probability)$ 
    \State $thread_{augmented} \leftarrow T$
    
    \For{$t_i$ in $tweet_{index}$}
        \State $thread_{augmented}[t_i] \leftarrow nlpaug( tweet[t_i] )$
    \EndFor
    \State $result \leftarrow result \cup thread_{augmented}$
\EndFor
\State RETURN result
\EndFunction
\end{algorithmic}
\end{algorithm}

\section{Proposed Model}\label{sec5}
In this study we focus on rumour detection on Twitter posts on different newsworthy events. For this purpose we propose two models MOS-GCN and MOS-GAT. In this section, we elaborate each step in our proposed method. We first explain the data preprocessing adopted in our method(section 5.1). Then we explain in detail the feature extraction process in section 5.2. Following this we explain two different model architectures used for detection rumors in section 5.3.

\subsection{Data preprocessing}\label{subsection51}
\begin{figure}
    \includegraphics[width=\textwidth]{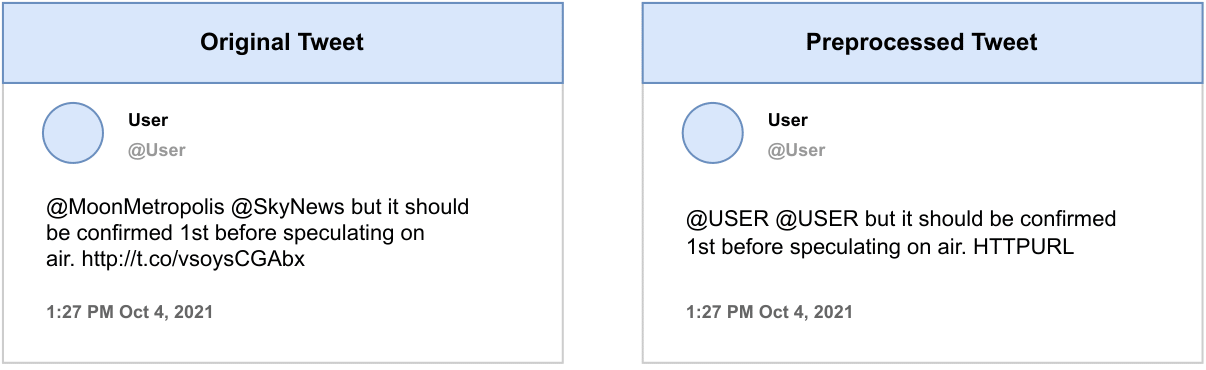}
    \centering
    \caption{Example of preprocessing and it's outcome.}
    \label{fig:preprocessedtweetexample}
\end{figure}

Twitter data contains significant amount of noise which can greatly hinder the performance of Transformer models\cite{https://doi.org/10.48550/arxiv.2204.09371, https://doi.org/10.48550/arxiv.2003.12932}, which can be reduced to large extent by pre-processing. 
To preprocess the data, firstly, we tokenzied the tweets using TweetTokenizer from the NLTK toolkit\cite{bird2009natural}. Secondly, the URLs and mentions are replaced with placeholders HTTPURL and @USER. Lastly, the emoji package\cite{jalilov_emoji_nodate} was used to translate emotion icons into text strings to capture the emotions signified by these emojis. Additionally, as we have restricted our domain specific to the English language, all non ASCII-English characters were removed from the tweets. An example of preprocessed tweet is shown Fig. \ref{fig:preprocessedtweetexample}.

\subsection{Feature extraction}\label{subsection333}
A tweet consists of multi-modal content(textual, audio, video and images), user information and tweet propagation information. Any combination of these can be used to represent a tweet. In abound of previous works it was observed that handcrafted representation has been largely used to represent the tweets.\cite{S_Punla2022-bl} In recent works, textual features were represented using Word2Vec\cite{word2vec} or GloVe\cite{glove} embeddings in combination with(out) handcrafted textual features.\cite{Wu2020-tw, S_Punla2022-pb, Rani2021-sd} We also observed that very few works focused on utilizing the user information which when combined with other features can greatly increase the proficiency of the models.\cite{Zubiaga2017ExploitingCF}

In view of the previous works, we propose two feature extractor that consists of Textual Feature Encoder and Social Feature Encoder. Textual Feature Encoder captures the evasive social media language. To further support the textual features we used social feature encoder to capture the social behaviour of users.

Our textual feature encoder uses  BERTweet for representing textual information. BERTweet uses BERT architecture which is a combination of stacks of transformer encoder and self-attention mechanism. One of the benefits of this architecture is the ability to capture long-term dependencies using bi-directional mechanism. The BERTweet was trained on over 850 million English tweets that enables this model to capture all the linguistic components of tweets. Along with linguistic components, BERTweet capture the emojis and correlates the information with the text.

To compliment the textual representations, we have used four handcrafted user behavioural information\cite{Zubiaga2017ExploitingCF}:
\begin{enumerate}
\item Tweet count: Number of tweets a user has posted on the platform. To normalize the count, a 10-base logarithm was considered after rounding up.
\item Listed count: Number of times other users on the platform added the user in a list. To normalize the count, a 10-base logarithm was considered after rounding up.
\item Follow ratio: It is a ratio of number of followers to number of following. To normalize the count, a 10-base logarithm was considered after rounding down.
\item Verified: A binary feature(true/false) to understand the verification procedure of the user.
\end{enumerate}

\subsection{Model Architecture}\label{subsection333}
In this section, we propose effective rumour detection models which effectively captures cues from responses, user behaviour and propagation structure. A tweet thread can be represented as a graph which is known as Propagation Structure. Propagation structure is a vital information as it captures the flow of information. We have considered top-down approach to represent a thread as it captures the flow of information from the source to the responses. In the real-world setting the natural flow of information is similar to the top-down approach. The propagation structure is initialized by a source tweet and each node is added to the existing structure when a user responds to a tweet creating a thread of responses. To effectively capture the propagation structure we use graph neural networks(GNN). Recently, GNN have achieved state-of-the-art results for link prediction, node prediction, etc., pertaining to graph structure.\cite{https://doi.org/10.48550/arxiv.1812.08434} We have used convolution-based GNN model(MOS-GCN) and attention-based GNN model(MOS-GAT) to enhance the classifier with rich representations of the propagation structure. 

A propagation structure is represented as adjacency matrix, \emph{A}(Equation \ref{equation:adjMatrix}). The adjacency matrix is a squared matrix with $N \times N$ size, where \emph{N} is the number of nodes in the graph. If an edge exists between any two node in the graph, $a \rightarrow b$, then the position (a,b) in the matrix is marked as 1. As we have taken directed graphs, $a \rightarrow b \neq b \rightarrow a$. In propagation graph all nodes have at least degree 1. 

\begin{equation}
 A = \begin{bmatrix}
   1 & 0 & 1 & \cdots & 1 \\
   0 & 0 & 1 & \cdots & 0 \\
   \vdots & \vdots & \vdots & \ddots & \vdots  \\
   0 & 1 & 0 & \cdots & 1 
 \end{bmatrix}
 \label{equation:adjMatrix}
\end{equation}

Each node in the graph can be represented by node feature. The aggregation of all node feature is called feature matrix(Equation \ref{equation:featureMatrix}), \emph{F} which is of size $N \times M$ where \emph{N} is the number of nodes in the graph and \emph{M} is the size of feature vector.

\begin{equation}
 F = \begin{bmatrix}
   f_1 & f_2 & \cdots & f_n
 \end{bmatrix}
 \label{equation:featureMatrix}
\end{equation}

The adjacency matrix along with feature matrix is used by MOS-GCN and MOS-GAT for rumour detection. In the below sections we explain in detail the \emph{Convolution-based GNN model(MOS-GCN)} and \emph{Attention-based GNN model(MOS-GAT)}.

\subsubsection{Convolution based Rumour Detection Model}\label{subsubsection333}
%
    
%
To embed the information of various interactions of users on a tweet we employ graph convolution neural network(GCN). Rumour detection model which deploys 
GCN captures spatial relationship among user comments and behaviours using adjacency matrix which represents the structural information of a Twitter thread. We also provide a rich node feature matrix, \emph{F}, to add tweet and user behavioural representations. 

To capture all representations the layer-wise propagation rule in Equation (\ref{equation:gcnEq}) is derived where $\hat{A}$ is the normalized adjacency matrix with self-loops obtained by $\hat{A} = A + I$ and $\hat{D}$ is the the diagonal node degree matrix of the normalized adjacency matrix.\cite{gcn} The feature matrix, \emph{F}, is used when l = 0(Equation (\ref{equation:gcnEq2})). The \emph{W} is the trainable weight matrix for the \emph{l}-th layer and $\sigma$ is the non-linear activation function(ReLU, GeLU, etc.).
\begin{equation}
H^{(l+1)} = \sigma(\hat{D}^{-1/2}\hat{A}\hat{D}^{-1/2}H^{(l)}W^{(l)})\textit{, where } \hat{A} = A + I
\label{equation:gcnEq}
\end{equation}
\begin{equation}
\textit{when l = 0, } H^{(0)} = \sigma(\hat{D}^{-1/2}\hat{A}\hat{D}^{-1/2}FW^{(0)})
\label{equation:gcnEq2}
\end{equation}

The tweets in the graph can use the layer-wise propagation rule to aggregate the neighbourhood representations of both textual and user behaviour embeddings(feature matrix, \emph{F}) to obtain tweet-level embeddings. These embeddings can be pooled to obtain Twitter thread embeddings which can be used to train multi-layer perceptron for rumour detection tasks.

\subsubsection{Attention based Rumour Detection Model}\label{subsubsection333}
A serious drawback of Convolution based Rumour Detection Model is that it assigns equal weights to all the neighbouring tweets.\cite{gat, Wu_2021} The tweets tend to not be of equal importance in terms of the content. 
Thus we utilize the attention mechanism to assign greater weights to supposedly more important tweets. 

Attention based Rumour Detection model learns relative weights between two connected response tweets. Initially the tweets are embedded into a higher levels and dense representations using Equation (\ref{equation:gatEq1}) where \emph{W} is the learnable weight matrix.\cite{gat} These tweet representations are then used to calculate pair-wise neighbouring tweets attention scores using Equation (\ref{equation:gatEq2}). The pair-wise tweets are first concatenated followed by dot product with learnable weight matrix \emph{a}, finally Leaky-ReLU is applied to obtain attention scores. To normalize the attention scores we apply Softmax to $e_{ij}$(Equation (\ref{equation:gatEq3})), enabling comparing of attention scores of different nodes possible. Finally, aggregation of all tweets in the Twitter thread is performed using attention scores as scaling parameter for the level of information each neighbouring tweets contribute(as shown in equation (\ref{equation:gatEq4})).
\begin{equation}
z^{(l)}_i = W^{(l)}H^{(l)}_i
\label{equation:gatEq1}
\end{equation}
\begin{equation}
e^{(l)}_{ij} = LeakyReLU( a^{(l)} \cdot ( z^{(l)}_i \parallel z^{(l)}_j ) )
\label{equation:gatEq2}
\end{equation}
\begin{equation}
\alpha^{(l)}_{ij} = \frac{\exp{e^{(l)}_{ij}}}{\sum_{j \in N_i}\exp{e^{(l)}_{ij}}}
\label{equation:gatEq3}
\end{equation}
\begin{equation}
H^{(l+1)} = \sigma(\sum_{j \in N_i} \alpha_{ij}^{k} z^{(l)}_j )
\label{equation:gatEq4}
\end{equation}
\begin{equation}
H^{(l+1)} = \parallel^K_{k=1}\sigma(\sum_{j \in N_i} \alpha_{ij}^{k} z^{(l)}_j )
\label{equation:gatEq5}
\end{equation}

Multi-head attention mechanism where the self-attention process is repeated for \emph{K} times is leveraged to stabilise the learning process(shown in Equation (\ref{equation:gatEq5})). Finally, feature-wise aggregation of all attention heads is performed to obtain tweet-level embeddings. 

\section{Experiment and Results}\label{sec6}
In this section, we conduct extensive experiments to verify performance of: (i) proposed oversampling technique, and (ii) proposed GNN models. For analysis we used two publicly available benchmark twitter datasets. We validated our results by comparing with multiple baselines. Additionally, we performed temporal early rumour detection. 

\subsection{Dataset}\label{subsection}
%
    
        

%

Experimental evaluation of our architecture was conducted on two publicly available benchmark datasets on Twitter, PHEME9 and PHEME5. PHEME9 dataset consists of more than 30,000 tweets which comprises of 9 individual newsworthy events. PHEME5 is a subset of of PHEME9 dataset, which comprises of 5 newsworthy events and more than 5000 number of Twitter threads. Each tweet thread in PHEME9 and PHEME5 were labelled as rumour(R)/non-rumour(NR). PHEME9 also consists three additional labels for rumourous tweet threads, which were true(T)/false(F)/unverified(U). Notably the dataset was extremely skewed which can be observed from Table \ref{table:dataDistribution}. For few events categories such as Ebola Essien, Gurlitt, there was absence of even single Twitter threads for specific labels. 


\begin{table}[ht]
\begin{tabular}{|c|ccc|cccc|ccc|c}
\hline
\multirow{2}{*}{
\parbox[c]{.2\linewidth}{\centering Events}}
  & \multicolumn{2}{c}{PHEME9} &&
\multicolumn{3}{c}{PHEME9} && \multicolumn{2}{c}{PHEME5} & \\ 
 & {R} & {NR} && {T} & {F} & {U} && {R} & {NR} &  \\ \hline
Charlie Hebdo & 458 & 1,621 && 193 & 116 & 149 && 458 & 1,621 &\\
Sydney siege & 522 & 699 && 382 & 86 & 54 && 522 & 699 &\\
Ferguson & 284 & 859 && 10 & 8 & 266 && 284 & 859 &\\
Ottawa shooting & 470 & 420 && 329 & 72 & 69 && 470 & 420 &\\
Germanwings crash & 238 & 231 && 94 & 111 & 33 && 238 & 231 &\\
Putin missing & 126 & 112 && 0 & 9 & 117 && - & - &\\
Prince Toronto & 229 & 4 && 0 & 222 & 7 && - & - & \\
Gurlitt & 61 & 77 && 59 & 0 & 2 && - & - &\\
Ebola Essien & 14 & 0 && 0 & 14 & 0 && - & - &\\ \hline
\end{tabular}
\caption{Data distribution for different datasets.} 
\label{table:dataDistribution}
\end{table}

\subsection{Baseline and experimental setup}
%

    
    

    
%

\subsubsection{Baselines}
For the purpose of validation of our proposed method, we compared it against following state-of-the-art baselines rumour detection systems. The systems are as follows:

\textbf{NileTMRG}: Battery of machine learning models(Logistic regression, Support Vector Machine, etc.) were trained on handcrafted features integrated with textual features generated via bag-of-words. Linear support vector machine(SVM) model outperformed other models for these features.

\textbf{BranchLSTM}: The Twitter thread's structural information is extracted by decomposing the propagation tree into branches. LSTM trained on the decomposed branches inherits the sequential nature of the tweets for classification. Handcrafted features were incorporated to enrich the tweet representation while training the model.

\textbf{RvNN}: In this technique, the propagation structure of tweet is captured by employing recursive neural networks. The tweets in the propagation structure are represented via word vectors computed using TF-IDF.

\textbf{Hierarchical GCN-RNN}: Conversations on a Twitter thread were captured by graph convolution networks combined with RNN to perform rumour detection and stance classification. 

\textbf{LSTM}: RNN combined with LSTM cells was trained with 50 dimension GloVe embedding for rumour detection.\cite{S_Punla2022-ic}

\textbf{CNN}: This technique applies convolution based method that used GloVe embedding with two layers of CNN for rumour detection.\cite{S_Punla2022-ic}

\textbf{CRF}: A sequential approach to learn dynamics of Twitter post by training a linear-chain CRF model which utilizes previous posts of a newsworthy events along with user comments.

\textbf{Han et. al.}: A data augmentation approach based on semantic relatedness. The augmented dataset was used to train a DNN model for rumour detection using DNN model.\cite{1907.07033}

\textbf{RDPNN}: A hybrid deep learning model was trained to learn individual tweet-level attributes along with user responses and tweet-level metadata. The model was trained using augmented PHEME5 dataset.\cite{2002.12683}

\subsubsection{Experimental settings}
Accuracy (Acc.) and micro-F1 score(F1) was used as a screening parameter to demonstrate the efficacy of our technique. Micro-F1 score was used as the distribution for the data was highly skewed.\cite{Zubiaga2017ExploitingCF}
In previous literature, evaluation using K-fold cross-validation has been prominent.\cite{Ma2021-ro, 10.1145/3477495.3531930, lin-etal-2021-rumor} However, on retrospective validation the performance of the model drops significantly over such evaluation techniques. As a result, we used Leave-One-Out Cross Validation(LOO-CV) accounting to a realistic scenario.\cite{Zubiaga2017ExploitingCF} In LOO-CV, each cross-validation set consists of \emph{n - 1} events in the training set, while leaving 1 event for the testing set. 

To further show the robustness of our proposed method we validated our method on two types of labels. The first one consists of \emph{Rumour} and \emph{Non-rumour} tweets which can be performed using PHEME5 and PHEME9 dataset. The second one consists of \emph{True}, \emph{False}, and \emph{Unverified} tweets which can be performed by using PHEME9 dataset. We implemented GNN-based approaches using Pytorch-Geometric library\cite{pytorch-geometric}, in addition to Pytorch framework\cite{pytorch}. The hyper-parameters for our proposed method are available in \ref{table:hp3label} and \ref{table:hp2label}.

\subsection{Results}
We perform comprehensive comparison of existing baselines with our architecture to validate efficacy of our method. We have illustrated the comparison of the models in Table \ref{table:pheme93label}, Table \ref{table:pheme92label} and Table \ref{table:pheme52label}. We saw a clear increase of more than 10\% across all the models which used the augmented dataset for training. Notable, we inferred that non-random data augmentation has lead to at least 7\% increase across both the GCN and GAT based architecture. This can be attributed to the inclusion criteria employed at the time of augmentation. Our method distills out contextually significant tweets in contrast to those which are less significant for augmentation. GAT with non-random data augmentation has outperformed all the models for PHEME9 dataset. For PHEME9 with three labels, NRA-GAT has achieved 79.22\% Accuracy and 48.88 F1-score while for PHEME9 with two labels, NRA-GAT has achieved 78.36\% Accuracy and 73.07 F1-score. GCN with non-random data augmentation has outperformed all the models for PHEME5 dataset with two labels by achieving 76.45\% Accuracy and 73.64 F1-score. In general GAT has outperformed GCN due to attention mechanism which assigns non-uniform importance to neighboring tweets in a thread. We inferred a clear trend of GCN and GAT outperforming baseline models even for models trained on non augmented datasets. We believe this may be due to feature extraction method where we have employed the state-of-the-art textual encoder along with handcrafted user feature encoder. We have demonstrated the generalizability of our method by validating the performance of our method on baseline baseline datasets such as PHEME9 and PHEME5.

\begin{table}[htbp]
\centering
\begin{tabular}{|ll|l|l|}
\hline &
Method & Acc. & F1 \\ \hline
& NileTMRG & 36.00 & 29.70\\
& BranchLSTM & 31.40 & 25.90\\
& RvNN & 34.10 & 26.40\\
& Hierarchical GCN-RNN & 35.60 & 31.70 \\  \hline
& MOS-GCN & 51.71 & 36.62 \\
& RA MOS-GCN & 64.05 & 42.29 \\
& NRA MOS-GCN & 68.08 & 45.57 \\ \hline
& MOS-GAT & 79.13 & 41.71 \\
& RA MOS-GAT & 79.13 & 43.04 \\
& \textbf{NRA MOS-GAT} & \textbf{79.22} & \textbf{48.88} \\ \hline
\end{tabular}
\label{table:pheme93label}
\caption{Experimental results on PHEME9 with True, False and Unverified labels}
\end{table}

\begin{table}[htbp]
\centering
\begin{tabular}{|ll|l|l|}
\hline
& Method & Acc. & F1\\ \hline
& LSTM & 58.27 & 52.17 \\
& CNN & 60.35 & 54.77 \\ \hline
& MOS-GCN & 74.24 & 65.71\\
& RA MOS-GCN & 75.88 & 69.76\\
& NRA MOS-GCN & 77.49 & 71.58\\ \hline
& MOS-GAT & 77.77 & 69.01\\
& RA MOS-GAT & 77.89  & 72.22\\
& NRA MOS-GAT & 78.36 & 73.07\\ \hline
\end{tabular}
\label{table:pheme92label}
\caption{Experimental results on PHEME9 with rumour and non-rumour labels}
\end{table}

\begin{table}[htbp]
\centering
\begin{tabular}{|ll|l|l|}
\hline
& Method & Acc. & F1\\ \hline
& CRF & - & 60.1\\
& Han et al. & 68.5 & 65.6\\
& RDPNN & 68.4 & 72.7\\ \hline
& MOS-GCN & 76.09 & 72.82\\
& RA MOS-GCN & 76.57 & 73.83\\
& NRA MOS-GCN & 77.52 & 74.51\\ \hline
& MOS-GAT & 74.96  & 70.60\\
& RA MOS-GAT & 75.70 & 72.31\\
& NRA MOS-GAT & 76.45 & 73.64\\ \hline
\end{tabular}
\label{table:pheme52label}
\caption{Experimental results on PHEME5 with rumour and non-rumour labels}
\end{table}

\vspace{5cm}

\subsection{Early Rumour Detection}

\begin{figure}[h]
\includegraphics[width=\textwidth]{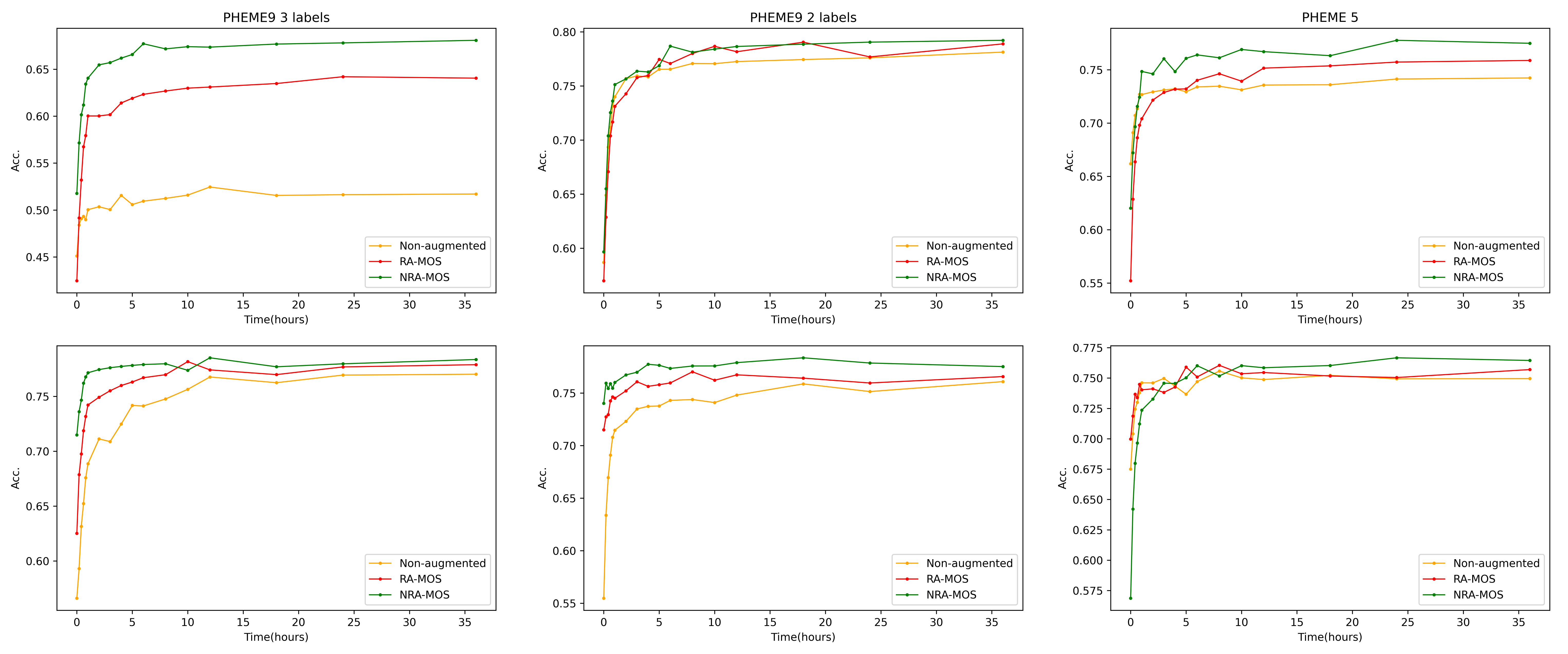}
\caption{Early rumour detection for PHEME9 and PHEME5 dataset.}
\label{figure:earlyrd}
\end{figure}

An ideal characteristic of rumour system is to screen out tweets as rumour during its advent. It is crucial to detect rumours at an early stage of propagation to mitigate any adverse impact caused by the rumourous tweet and prevent any further spread. In early rumour detection task we evaluate the performance of detection methods across series of checkpoints referred as 'delay'. The checkpoints are conditioned based on the time elapsed since the source tweet is posted.

To evaluate the performance of a rumour detection system across temporal resolution, we created multiple cohorts of tweets across varying checkpoints. For a twitter thread, the source tweet is considered as the epicentre of propagation graph. The time of arrival of the source tweet is considered as the $0^{th}$ hour, with respect to which the time of elapse for the other comments are calculated. Cohort of tweets across multiple threshold time-points are generated where-in all tweets preceding the cutoff time are disjointed(eliminated) from the thread following the source tweet. $N^{th}$-hour cohort is defined as cohort inclusive of the source tweet along with all the comments which were tweeted before Nth-time. Furthermore, we applied leave-one-out cross-validation technique for evaluating the performance of our architecture for the early rumour detection task.

Figure \ref{figure:earlyrd} shows the performance of our methods across all cohorts with 17 selected checkpoints for PHEME9 three-labels, PHEME9 two-labels and PHEME5 two-labels respectively. 

From Figure \ref{figure:earlyrd}, we observe that with increase in  the number of elapsed hours, the detection accuracy increases, which can be linearly correlated to the increase in the propagation information. We believe that the comments following the source tweets substantiates proof for the source tweet to be rumour. We infer a linear increment in the accuracy within the 4 hours after the source tweet is posted which tends to flatten after a lapse of time. The flattening of curve can be attributed as less information gain across the source tweet after a definite interval. For the all the rumour detection methods, we observe that the early performance fluctuates with time, hence not providing a smoothing curve. This is attributed to the fact that newer responses might be inconclusive in nature which can attribute as noise for these models. We also observed that the oversampling technique not only improves the accuracy of the model but also improves the early detection performance of the model. This also concludes to the robustness of our model across multiple check points while detecting rumour at an early stage. In general, models trained using non random augmentation performs better than any other technique for early rumour detection.

\section{Conclusion}\label{sec7}
Majority literature for rumour detection has subsided to techniques which has not implemented approaches for augmenting or synthesizing the datasets. In this paper, we propose a novel oversampling method MOS which synthesizes unique samples form the  existing oblique benchmark dataset, while inheriting the the contextual dependencies of a tweet. The augmented samples propagates linguistic characteristics in existing dataset contributing in diversifying and improving quality of the dataset. We trained our models over the augmented dataset which enhanced the performance illustrating  generalazbility and robustness of our method. Possibly, we believe augmenting the dataset had reduced overfitting of our method on the unseen test samples which enhanced the efficacy. Furthermore, we propose two GNN models(MOS-GCN and MOS-GAT) to extract critical information from user responses and user behavior of the Twitter thread network for rumour detection. Results on two benchmark rumour detection dataset have shown: (i) GNN models in conjunction with oversampled dataset vastly improve upon the existing state-of-the-art baseline classifiers; and (ii) potential of non-random selection criteria in oversampling for rumour detection.

Futhermore, in future we plan to: (i) Build an end-to-end model that incorporates both graph neural network and BERT for the classification task. (ii) Explore the biases introduced by model, hence mitigating such biases to build more robust techniques.

\section*{Acknowledgments}
We would like to thank Ridam Pal for his valuable suggestions.

\bibliographystyle{elsarticle-num} 
\bibliography{main}

\appendix
\section{Tweet relevancy}
\begin{figure}[H]
    \centering
    \includegraphics[width=\textwidth]{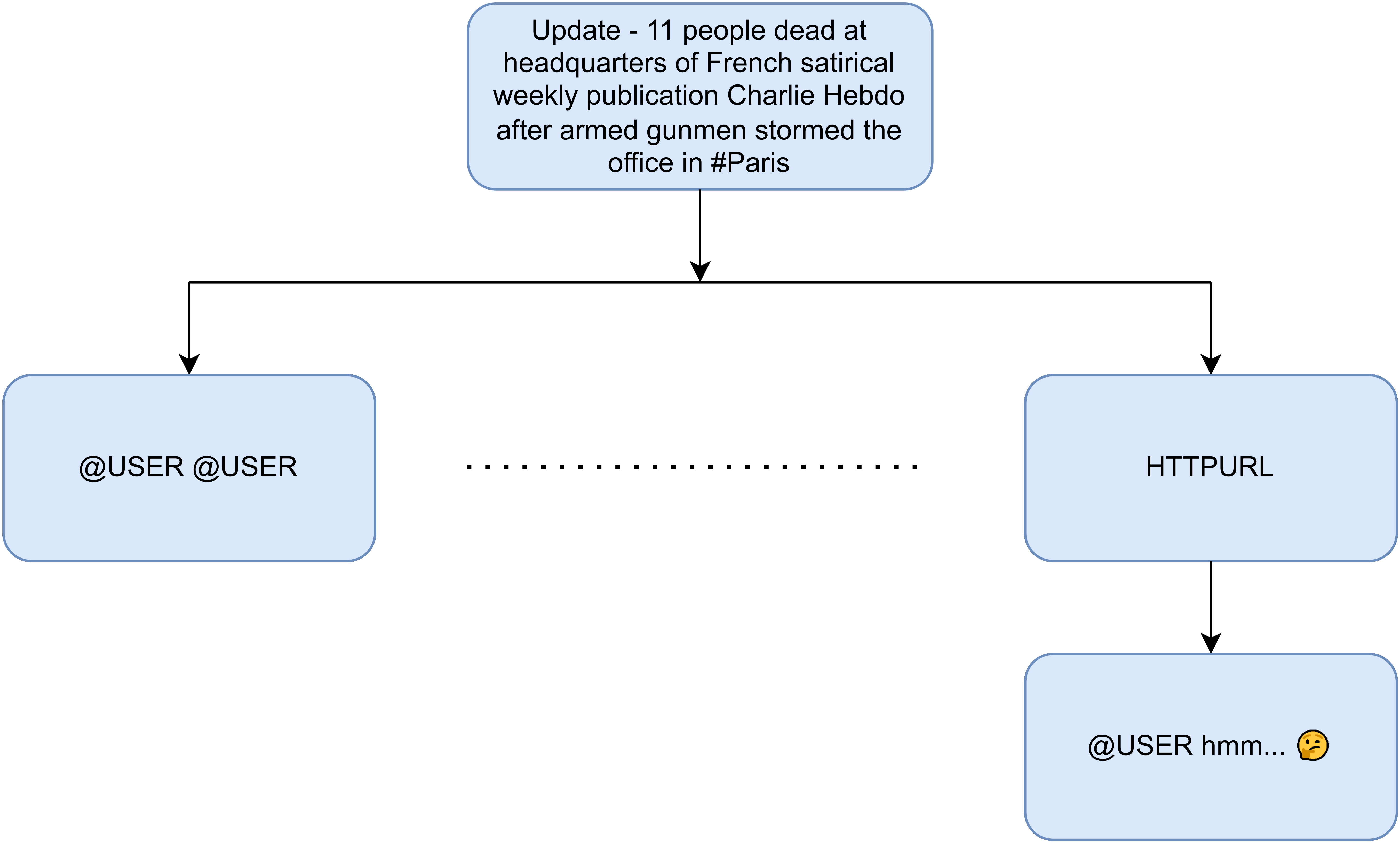}
    \caption{An example of Twitter thread to explain the importance for non-random selection criteria.}
    \label{fig:relevancy}
\end{figure}

\vspace{5cm}

\section{Hyper-paramters}
\begin{table}[htbp]
\centering
\begin{tabular}{|ll|l|}
\hline &
Hyper-parameter & Value \\ \hline
& Learning rate & $5\times10^{-3}$ \\
& Weight decay & $1\times10^{-3}$ \\
& Batch size & 128 \\
& Epochs & 100 \\
\hline
\end{tabular}
\label{table:hp3label}
\caption{Hyper-parameters for training GNN models on PHEME9 with 3 labels.}
\end{table}

\begin{table}[htbp]
\centering
\begin{tabular}{|ll|l|}
\hline &
Hyper-parameter & Value \\ \hline
& Learning rate & $5\times10^{-3}$ \\
& Weight decay & $1\times10^{-3}$ \\
& Batch size & 128 \\
& Epochs & 100 \\
\hline
\end{tabular}
\label{table:hp2label}
\caption{Hyper-parameters for training GNN models on PHEME9 with 2 labels and PHEME5.}
\end{table}

\end{document}